\documentclass[final]{cvpr}

\usepackage{times}
\usepackage{epsfig}
\usepackage{graphicx}
\usepackage{amsmath}
\usepackage{amssymb}
\usepackage{mathtools}
\usepackage{amsmath,amssymb,amsfonts}
\usepackage{amsthm}
\usepackage{titling}
\usepackage{multicol}
\usepackage{graphicx}
\usepackage{textcomp}
\usepackage{xcolor}
\usepackage{bbm}
\usepackage{algorithmic}
\usepackage{algorithm}
\usepackage{nicefrac}
\usepackage[pagebackref=true,breaklinks=true,colorlinks,bookmarks=false]{hyperref}

\def\cvprPaperID{****} % *** Enter the CVPR Paper ID here
\def\confYear{CVPR 2021}
\begin{document}

%%%%%%%%% TITLE
\title{Dynamic VAEs with Generative Replay for Continual Zero-shot Learning}

\author{Subhankar Ghosh\\
Indian Institute of Science\\
%Bengaluru, India, 
{\tt\small subhankarg@alum.iisc.ac.in}}
\date{}
\maketitle

%%%%%%%%% ABSTRACT
\begin{abstract}
	Continual zero-shot learning(CZSL) is a new domain to classify objects sequentially the model has not seen during training. It is more suitable than zero-shot and continual learning approaches in real-case scenarios when data may come continually with only attributes for a few classes and attributes and features for other classes. Continual learning(CL) suffers from catastrophic forgetting, and zero-shot learning(ZSL) models cannot classify objects like state-of-the-art supervised classifiers due to lack of actual data(or features) during training. This paper proposes a novel continual zero-shot learning (DVGR-CZSL) model that grows in size with each task and uses generative replay to update itself with previously learned classes to avoid forgetting. We demonstrate our hybrid model(DVGR-CZSL) outperforms the baselines and is effective on several datasets, i.e., CUB, AWA1, AWA2, and aPY. We show our method is superior in task sequentially learning with ZSL(Zero-Shot Learning). We also discuss our results on the SUN dataset. Our code is available at {\url{https://github.com/DVGR-CZSL/DVGR-CZSL.git}}.
	
\end{abstract}
\section{Introduction}

Although artificial neural networks(ANNs) show excellent performance on many machine learning problems such as classification, object detection, and natural language processing, they forget the previous knowledge when trained with new tasks. The conventional deep learning algorithms heavily depend on large labeled data, but nowadays, it is impractical to annotate everything surrounding us. This leads us to think of a method that can learn sequentially and simultaneously works for zero-shot learning(ZSL). 

Many researchers over the years have addressed the ZSL\cite{a1, a2, a3} and CL\cite{a4, a5, a6} problems separately, while few worked on continual zero-shot learning(CZSL)\cite{a7, a8, a9, a10}. ZSL is a problem in machine learning, where at test time, a learner observes samples from classes that are not observed during training and needs to predict the class they belong to. There are various methods to tackle the ZSL, but they are still unable to learn continually from a sequence of tasks without forgetting previously learned knowledge. Acquiring new knowledge and leveraging past experience from streaming of data is termed continual learning(CL). This strategy needs to be merged with ZSL to enable continual zero-shot learning. 

\begin{figure}[hbt!]
	\centering
	\includegraphics[width = 8cm, height = 6cm]{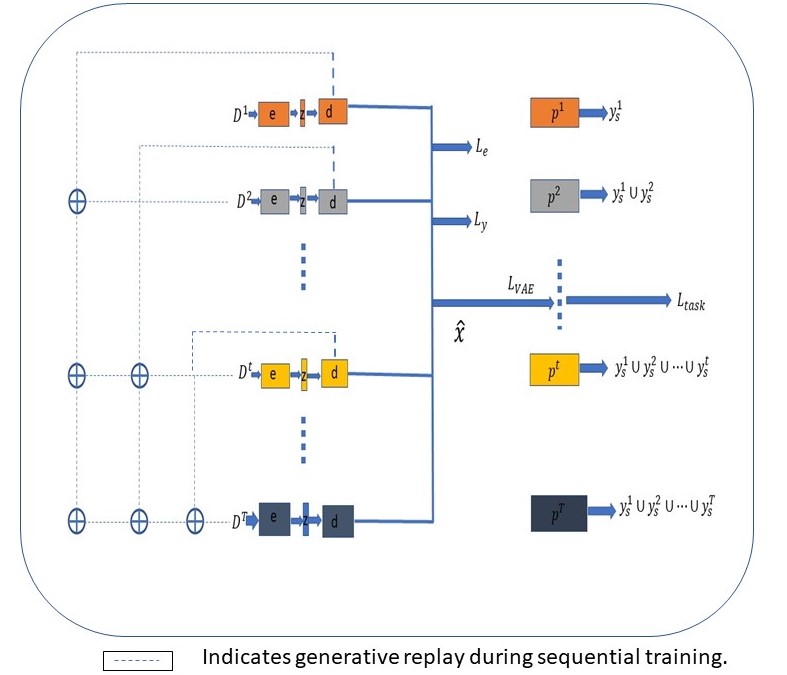}
	\caption{is our model at training time. Architecture growth occurs at the arrival of $t^{th}$ task by adding a conditional VAE, and a ANN denoted as $p^t$. To prevent forgetting, private VAEs are stored for each task. The first private VAE gets trained using real data from the first task. The second private VAE is trained using real data from the second task and synthesized data corresponding to the first task's classes generated from the first private decoder. Similarly, the $t^{th}$  private VAE sees real data from the $t^{th}$ task, synthesized data from the previous private decoders corresponding to their tasks' classes during training. Each color represents an individual variational autoencoder(VAE). (+) sign indicates concatenation. 
	}
	\label{figure:f1}
	
\end{figure}

\begin{figure}
	\centering
	\includegraphics[width = 8cm, height = 5cm]{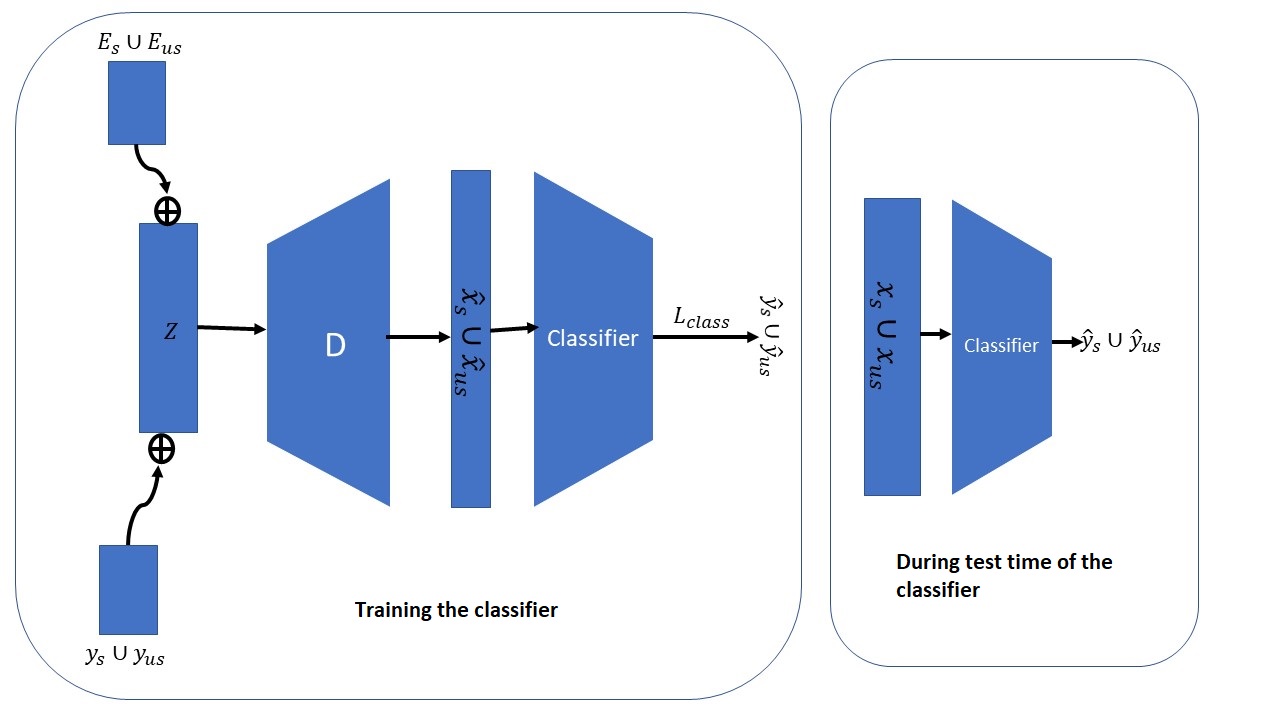}
	\caption{ (Left) is our model at the $t^{th}$ task during synthesizing data for seen and unseen classes that further are sent to a classifier for training the classifier. (Right) is the classifier during inference. (+) sign indicates concatenation. $\mathcal{Y}$ and E indicate class labels and embeddings, respectively. $\mathcal{X}$ and $\hat{\mathcal{X}}$ indicate actual and synthesized features, respectively.
	}
	\label{figure:f2}
\end{figure}

A ZSL model recognizes unseen classes through transferring knowledge from seen to unseen classes through class embeddings\cite{a11}. Despite showing promising performance, the ZSL models are unable to learn from a sequence of tasks. Generative approaches have received quite an attention over embedding approaches due to the ability to generate synthetic data of unseen classes\cite{a12, a13, a14} using only class attributes. We propose a novel hybrid model that consists of  T conditional variational autoencoders(VAEs)\cite{a15}. The model uses generative and architecture-based approaches to alleviate forgetting. The first conditional VAE transfers knowledge from seen to unseen classes for the first task, the second conditional VAE does the same for the classes of the first and second tasks, and similarly, the $T^{th}$ conditional VAE transfers knowledge from seen to unseen classes for all T tasks(T is the number of tasks the model needs to learn). Our idea is motivated by the fact that processing and synthesizing images are time taking for CL when the number of classes is high. Instead of images, we use features of benchmark datasets generated from a pre-trained model to train and test our proposed method. The architecture of our model is given in Figure \ref{figure:f1} and Figure \ref{figure:f2} for the training and testing, respectively. The main contributions of this work are summarised as follows:
\begin{itemize}
	\item We propose a novel generative replay-based and structure-based approach that can identify unseen classes through class embeddings' information using conditional VAEs.
	\item The proposed model is developed for a single head setting, and it does not require any task information at the inference time. Therefore, the model is more convenient to solve real-case problems.
	\item We present results on five ZSL benchmark datasets and show that our model achieves state-of-the-art performance on the four of the datasets.
\end{itemize}

\section{Related Work}
\subsection{Zero-shot Learning(ZSL)}
ZSL was first introduced at\cite{a16} for attribute-based classification and considered a disjoint setting. The ZSL\cite{a1, a2, a3, a34} has recently attracted much attention due to its ability to classify objects of new classes by transferring knowledge from seen to unseen labels through a semantic relationship between classes and their attributes. We transform a ZSL problem into a supervised classification problem using generative models trained with seen class data. Once they can generate data for seen classes, they can synthesize data for unseen classes.
\subsection{Continual Learning(CL)}
There are three types of continual learning approaches: architecture-based, regularization-based, and rehearsal-based.
\subsection*{Architecture-based Methods}
This is the first approach to prevent catastrophic forgetting, and it changes the network's structure by growing a module for each task either physically or logically\cite{a17, a18}. Previous tasks' performance is maintained by storing the learned modules while accommodating new tasks by expanding the network with new modules. PNNs\cite{a19} statically grow the architecture and are immune to forgetting. They utilize prior knowledge via lateral connections to previously learned features. Reference\cite{a20} proposed a dynamically expandable network(DEN) that can continually decide its network's capacity as it gets trained on a sequence of tasks. The computational cost is inevitable if we work with mentioned methods, where many tasks need to be learned, and fixed capacity memory can not be granted. 
\subsection*{Regularization-based Methods}
The second family of this field is regularization-based methods\cite{a21, a22, a45, a46, a23}. They evaluate the importance of a network's parameters and penalize those weights while switching from one to another task. Few approaches are there for penalizing the weights; one of them is the elastic weight consolidation[EWC]\cite{a21}, where important parameters have the highest in terms of the Fisher information matrix. In reference\cite{a22}, the weights are kept track of how much the loss changes due to change in specific weights and include this information during continual training. Reference\cite{a23} centers on the change on the activation instead of considering the loss's change. This way, parameter importance is learned in an unsupervised manner. The number of tasks often limits these methods.
\subsection*{Rehearsal-based Methods}
The last family of methods of this domain to alleviate forgetting is rehearsal-based. Existing methods use two approaches: either store a few samples per class from previous tasks or train generative models to sample synthetic data from earlier learned distributions. The iCaRL\cite{a24} stores a subset of real data(exemplars). For a fixed memory resource, the number of data stored per learned class reduces as the number of tasks rises, so the models' performance decay. References\cite{a25, a26} present a bias-correction layer to correct the original fully-connected layer's output to address the data asymmetry between the old and new classes. A few studies on tiny episodic memories in continual learning are GEM\cite{a27}, A-GEM\cite{a28}, MER\cite{a29}, and ER-RES\cite{a30}. 

The second method in this family does not store any data but synthesizes data using generative models. Reference\cite{a31} used generative replay with an unconditional GAN, where an auxiliary classifier requires to determine which classes the generated samples belong to. Reference\cite{a32} is an improved version of\cite{a31}, where they used class-conditional GAN to generate data. Reference\cite{a33} used a generative autoencoder for replay. Synthetic data for prior tasks are generated based on the mean and covariance matrix using the encoder's class statistics. The assumption of a Gaussian distribution of the data is the major limitation of these approaches.
\subsection{Continual Zero-shot Learning(CZSL)}
Reference \cite{a7} proposed a continual zero-shot learning model called GCZSL, a single head CZSL where the task identity is revealed during training but not during testing. To mitigate forgetting, they used knowledge distillation and stored few real data per class using a small episodic memory.  Reference\cite{a8} proposed a CZSL model called GRCZSL that handles catastrophic forgetting by replaying synthetic samples of seen classes. The replay samples are generated using trained conditional VAE over the immediate past task. Reference\cite{a9} proposed a CZSL model that has two parts: shared and private. The shared module gets trained for all tasks, whereas the private modules are task-specific grows with each task. The model uses generative replay from the decoder of shared module and architecture growth to prevent catastrophic forgetting. The methods mentioned above and\cite{a10} are for the single-head setting. There are a few works\cite{a35, a36} for the multi-head setting. Reference\cite{a35} proposed an average gradient episodic memory(A-GEM) based CZSL, and\cite{a36} offered a generative model-based CZSL.
\section{Dynamic VAEs with Generative Replay for Continual Zero-shot Learning}
We study the problem of learning a sequence of T data distributions denoted as $D = \{D^1, D^2, ..., D^T\}$, where $D^t = \{(X^t_i, Y^t_i, E^t_i, T^t_i)^{n_t}_{i = 1}\}$ is the data distribution for the task t with $n_t$ sample tuples of input($X^t \in \mathcal{X}$), target label ($Y^t \in \mathcal{Y}$), class embeddings($E^t \in \mathcal{E}$), and task label($T^t \in \mathcal{T}$). The goal is to learn a sequential function, $f_\theta: D^t \rightarrow \hat{\mathcal{Y}}^t$, for each task, where $\hat{\mathcal{Y}}^t$ are the predicted labels corresponding to $t^{th}$ task. $f_\theta \in (f_P \cup f_p)$, where $f_P:D^t \rightarrow (\hat{\mathcal{X}^t}, \hat{\mathcal{Y}^t}, \hat{\mathcal{E}}^t)$, and $f_p:\hat{\mathcal{X}^t} \rightarrow \hat{\mathcal{Y}}^t$\footnote{{$\hat{}$} indicates synthesized data.}. We try to achieve our goal by training a module for each task with a generative replay to mitigate catastrophic forgetting of prior knowledge. The model prevents catastrophic forgetting and begins learning $f_{\theta}^t$ where $\theta \in (\theta_P, \theta_p)$ as mapping function from ${D}^t$ to $\mathcal{Y}^t$. We use some n samples per class to be synthesized prior to $t^{th}$ task and accumulate the generated data to the current task($t^{th}$) to train the model. During training the model with $t^{th}$ task:
\begin{center}
	$X^{t} \leftarrow X^{t}\cup \hat{\mathcal{X}}^{1:(t-1)}$
\end{center}
The cross-entropy loss function for the $f^t_{\theta}$ mapping corresponds to:
\small
\begin{equation}
	\begin{split}
		L_{task}(f^t_{\theta}, D^t) = -\mathop{\mathbb{E}}_{(\mathcal{X}^t, \mathcal{Y}^t, \mathcal{E}^t, \mathcal{T}^t) \sim D^t}\\
		\left[\sum_{c = 1}^C\mathbbm{1}_{(c = \mathcal{Y}^t)}log(\sigma(f_{\theta}^t(\mathcal{X}^t, \mathcal{Y}^t, \mathcal{E}^t, \mathcal{T}^t)))\right]
	\end{split}	
\end{equation}
\normalsize
Where $\sigma$ is the softmax function, in learning a sequence of tasks, an ideal $f^t_{\theta}$ maps the input images $X^t$ to their predicted labels $\hat{\mathcal{Y}}^t$.
\subsection*{Variational Autoencoders(VAEs)}
Autoencoders can efficiently learn input space and representation\cite{a37, a38}. A VAE follows an encoder-latent vector-decode architecture of classical autoencoder, which places a prior distribution on the input space and uses an expected lower bound to optimize the learned posterior. Conditional VAE is an improved version of the VAE, where data are fed to network with class properties such as labels or embeddings, or both. The conditional VAE is the fundamental building block of our approach. Variational distribution tries to find a true conditional probability distribution over the latent space z by minimizing their distance using a variational lower bound limit.The loss function for a VAE is:
\begin{multline}
	L_{VAE} = \mathop{\mathbb{E}}_{q_{\phi}({z}|{x})}\left[log(p_\theta({x}|{z}))\right] - D_{KL}(q_\phi({z}|{x})\parallel p_{\theta(z)})
\end{multline}
Where the first term is the reconstruction loss, and the second one is the KL divergence between $q({z}|{x})$ and p(z). The encoder predicts $\mu$ and $\sum$ such that $q_{\phi}({z}|{x}) = \mathcal{N}(\mu, \sum)$, from which a latent vector is synthesized via reparametrization process.
\subsection*{Other Losses}
The mean square loss function for the $f_P^t$ mapping corresponds to:
\begin{multline}
	L_y(f^t_{P}, D^t) = \mathop{\mathbb{E}}_{\hat{\mathcal{Y}}}
	\left[\hat{\mathcal{Y}} - \mathcal{Y}\right]^2
\end{multline}
$L_y$ is the mean square loss between actual labels and predicted labels for the $t^{th}$ task.
\begin{multline}
	L_e(f^t_{P}, D^t) = \mathop{\mathbb{E}}_{\hat{\mathcal{E}}}
	\left[\hat{\mathcal{E}} - \mathcal{E}\right]^2
\end{multline}
$L_e$ is the mean square loss between actual class embeddings and predicted class embeddings for the $t^{th}$ task.
The final objective function of our method for the $t^{th}$ task is: %tth
\begin{multline}
	L^{(t)} = \lambda_1 L_{task} + \lambda_2 L_{VAE} + \lambda_3 L_{y} + \lambda_4 L_{e}
\end{multline}
Where, $\lambda_1, \lambda_2, \lambda_3$ and $\lambda_4$ are regularizer constants to control the effect of each loss component. The working algorithm of this model is presented in Algorithm \ref{alg:algorithm}. 
\subsection{Avoid forgetting}
Catastrophic forgetting occurs because of the asymmetry in the data between previous and new labels that creates a bias\cite{a47} in the network towards the current ones during training, and the NN models almost forget previous knowledge. Our approach's insight is to use generative replay with architectural growth to prevent forgetting. The DVGR-CZSL gets trained in the following way: The first module sees the first task's real data; the second conditional VAE sees real data of the second task and synthetic data of the first task. Similarly, the third conditional  VAE gets trained with real data of the third task and synthetic data of the first and second tasks generated from the first and second private decoders, respectively. It continues till the $T^{th}$ task.
\begin{algorithm}
	\caption{Continual Zero-shot Learning}
	\label{alg:algorithm}
	\textbf{Input}: $(\mathcal{X}, \mathcal{Y}, \mathcal{E}, \mathcal{T}) \sim D^{all}$\\
	\textbf{Parameters}: $\theta_P, \theta_p, \theta_c$\\
	\textbf{Output}: $\hat{\mathcal{Y}}$
	\begin{algorithmic}[1] %[1] enables line numbers
		
		\STATE $D^{gen} \leftarrow \{\}$
		\FOR{t $\leftarrow$ 1 to T}
		\FOR{e $\leftarrow$ 1 to epochs}
		\STATE Compute $L_{task}$ using $(\mathcal{X}^t, \mathcal{Y}^t, \mathcal{E}^t) \in D^t$ 
		
		\STATE Compute $L_{VAE}$ using $(\mathcal{X}^t, \mathcal{Y}^t, \mathcal{E}^t)\in D^t$
		\STATE Compute $L_y$ using $\hat{\mathcal{Y}}$ and $\mathcal{Y}$ 
		\STATE Compute $L_e$ using $\hat{\mathcal{E}}$ and $\mathcal{E}$
		
		\STATE $L^{(t)} = \lambda_1 L_{task} + \lambda_2 L_{VAE} + \lambda_3 L_{y} + \lambda_4 L_{e}$
		\STATE $\theta^{'} \leftarrow \theta - \alpha \nabla L^{(t)}$
		
		\ENDFOR
		\STATE Generate data from the trained module for seen and unseen classes to train a separate classifier.
		\STATE $D \leftarrow D_{s} \cup D_{us}$
		\FOR{$C_e \leftarrow$ 1 to $C_{epochs}$}
		\STATE Compute $L_{class}$ using ($\hat{\mathcal{X}}, \mathcal{Y}$) $\in$ D
		\STATE $\theta_{c}^{'} \leftarrow \theta_c - \alpha_c \nabla L_{class}$
		\ENDFOR
		\STATE Test the classifier for seen data. 
		\STATE Test the classifier for unseen data(ZSL).
		\FOR{c $\leftarrow$ 1 to C} 
		\STATE C is the replay classes.
		\FOR{i $\leftarrow$ 1 to n} 
		\STATE n is the number of samples to be generated per class for the generative replay.
		\STATE $ D^{gen} \in (\hat{\mathcal{X}}_i, \mathcal{Y}_i, \mathcal{E}_i)$
		
		\ENDFOR
		\ENDFOR
		\STATE $D^{t+1} \leftarrow D^{t+1}\cup D^{gen}$
		
		\ENDFOR
	\end{algorithmic}
\end{algorithm}
\subsection{Datasets}
We evaluate our model on five benchmark datasets used for ZSL: Attribute Pascal and Yahoo(aPY)\cite{a40}, Animals With Attributes(AWA1, AWA2)\cite{a39}, Caltech-UCSD-Birds 200-2011(CUB)\cite{a42}, and SUN\cite{a41}. The details of the datasets are presented in Table \ref{table:t1}.
\begin{table*}
	\centering
	\begin{tabular}{ |p{2cm}|p{1.5cm}|p{1.8cm}|p{0.7cm}|p{0.7cm}|p{4.5cm}|  }
		\hline
		Dataset & Size & Granularity & $\mathcal{E}$ & $\mathcal{Y}$ & \#images(Training + Testing)\\
		\hline
		SUN\cite{a41}   & medium &fine   &102&   708&11328 + 2832 \\
		\hline
		CUB\cite{a42} & medium & fine & 312 & 200 & 9440 + 2348\\
		\hline
		AWA1\cite{a39} & medium & coarse & 85 & 50 & 24382 + 6093\\
		\hline
		AWA2\cite{a39} & medium & coarse & 85 & 50 & 29860 + 7462\\
		\hline
		aPY\cite{a40} & small & coarse & 64 & 32 & 12272 + 3067\\
		\hline
	\end{tabular}
	\caption{Statistics for SUN\cite{a41}, CUB\cite{a42}, AWA1\cite{a39}, AWA2\cite{a39}, and aPY\cite{a40} are presented in terms of size, granularity, number of attributes($\mathcal{E}$), number of classes($\mathcal{Y}$), and number of images.}
	\label{table:t1}
\end{table*}
\subsection{CZSL setting}
The dataset we used in all our experiments follows the setting used in\cite{a43}. Classes that are seen or unseen are decided based on the model that has been trained so far with the classes. If a model goes trained continually up to the $t^{th}$ task, the classes are assumed to be seen till the $t^{th}$ task, and the rest of the classes are accepted unseen for the model while training. 
\subsection{Evaluation Matrices}
We evaluate our model on all tasks similar to \cite{a10} after training for each task. The following evaluation matrices are used for continual zero-shot learning after the $t^{th}$ task:
\begin{itemize}
	\item \textbf{Mean Seen Accuracy(mSA)}. We compute CZSL-S after tasks $t = 1, 2, 3, ... , T$ and take the average:
	\begin{multline}
		mSA = \frac{1}{T}\sum^{T}_{t = 1}ACC_S(D_{t_{s}}^{\leq{t}}, E^{\leq{t}})
	\end{multline}
	\item \textbf{Mean Unseen Accuracy(mUA)}. We compute CZSL-U after tasks $t = 1, 2, 3, ..., T-1$(we do not compute it after task T-1, since $D^{>{T -1}}_{us} = \phi$) and take the average: 
	\begin{multline}
		mUA = \frac{1}{T-1}\sum^{T-1}_{t = 1}ACC_U(D_{t_{s}}^{>{t}}, E^{>{t}})
	\end{multline}
	\item \textbf{Mean Harmonic Accuracy(mH)}. We calculate CZSL-H after tasks $t = 1, 2, 3, ..., T-1$ and take the average:
	\begin{multline}
		mH = \frac{1}{T-1}\sum^{T-1}_{t = 1}ACC_H(D_{t_{s}}^{\leq{t}},D_{t_{s}}^{>{t}}, E)
	\end{multline}
\end{itemize}
\section{Experiments}
This section consists of baselines and implementation details we used in our experiment.
\subsection{Baselines}
A handful of works are there previously done for continual zero-shot learning. References\cite{a7, a8, a9, a10} worked before on CZSL for single head setting, and our work is based on the same setting. 
\begin{itemize}
	\item \textbf{AGEM + CZSL\cite{a10}:} It is an average gradient episodic memory-based CZSL. The harmonic mean for the CUB and SUN datasets is presented there.
	
	\item \textbf{SEQ + CVAE\cite{a7}:} The authors trained conditional VAE sequentially without taking any CL strategy into account.
	
	\item \textbf{SEQ + CADA\cite{a7}:} The authors have trained CADA sequentially to develop a baseline without considering any CL strategy.
\end{itemize}
\subsection{Implementation Details}
We take PyTorch as our working framework. We train our proposed model for 101 epochs and the classifier for 25 epochs for each task. We perform experiments using 50 synthetic samples per class as the generative replay during continual training. To train the classifier, we took 150 samples per class for both seen and unseen classes. The Adam optimizer\cite{a44} has been used for all experiments, and the learning rate for the model and the classifier are  0.0001. The dimension of the latent variables z has been taken 50 for all our experiments. We take $\lambda_1 = \lambda_2 = \lambda_3 = \lambda_4 = 1$ at the loss function. We use a Tesla V100 GPU in all our experiments.
\section{Results and Discussion}
In the first set of experiments, we measure mSA, mUA, mH, the memory used by our method, and the size of our model's parameters and compare it against state-of-the-art methods on SUN, CUB, aPY, AWA1, and AWA2 datasets. We compare the harmonic mean accuracies for all methods in Table \ref{table:t7}.
\subsection{Performance on SUN Dataset}

\begin{figure}
	\includegraphics[width = 8cm, height = 5cm]{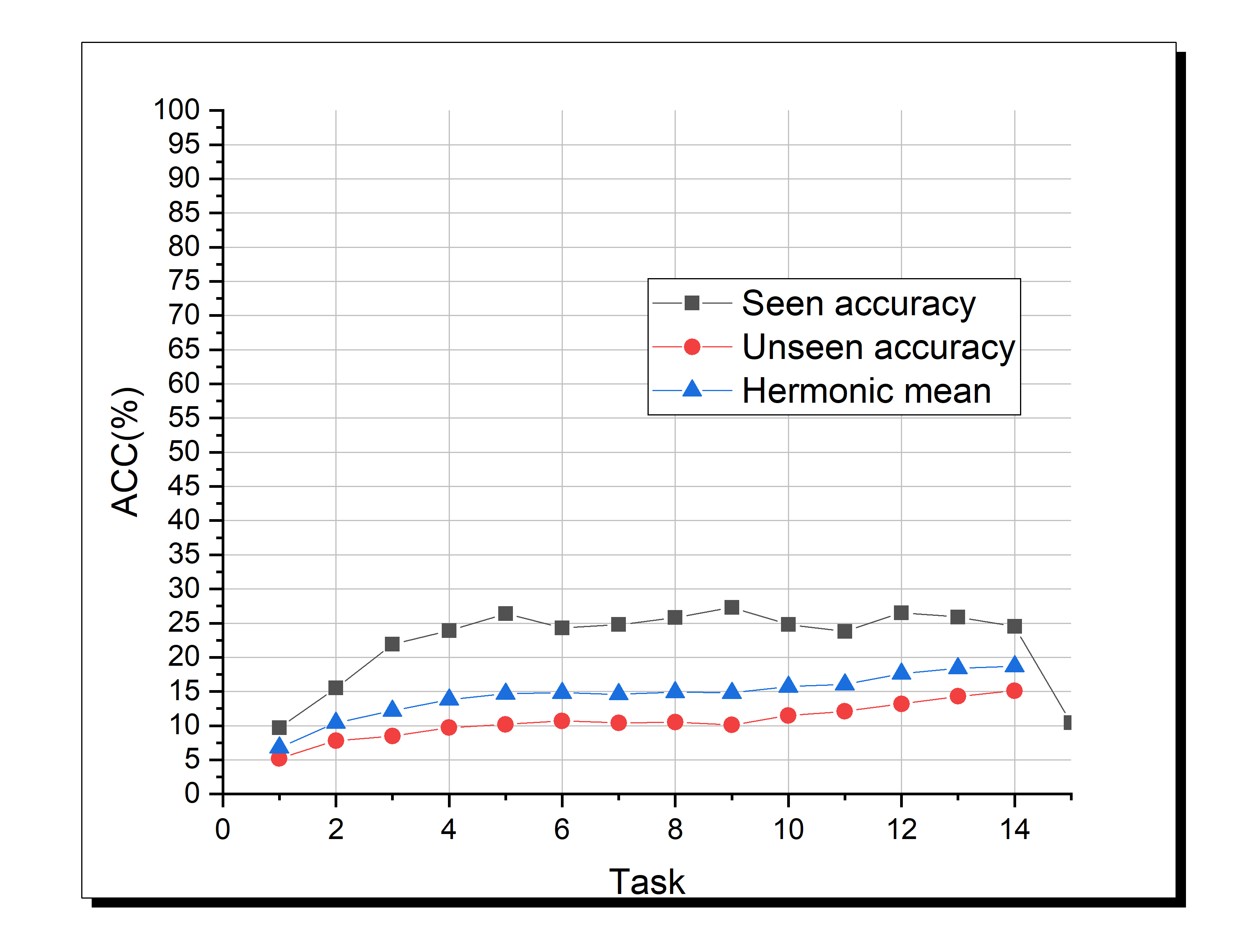}
	\caption{The SUN dataset results when we use the additional losses($L_e$ and $L_y$) during training the model.}
	\label{fig:f3}
	%    \vspace{<1>}
	\includegraphics[width = 8cm, height = 5cm]{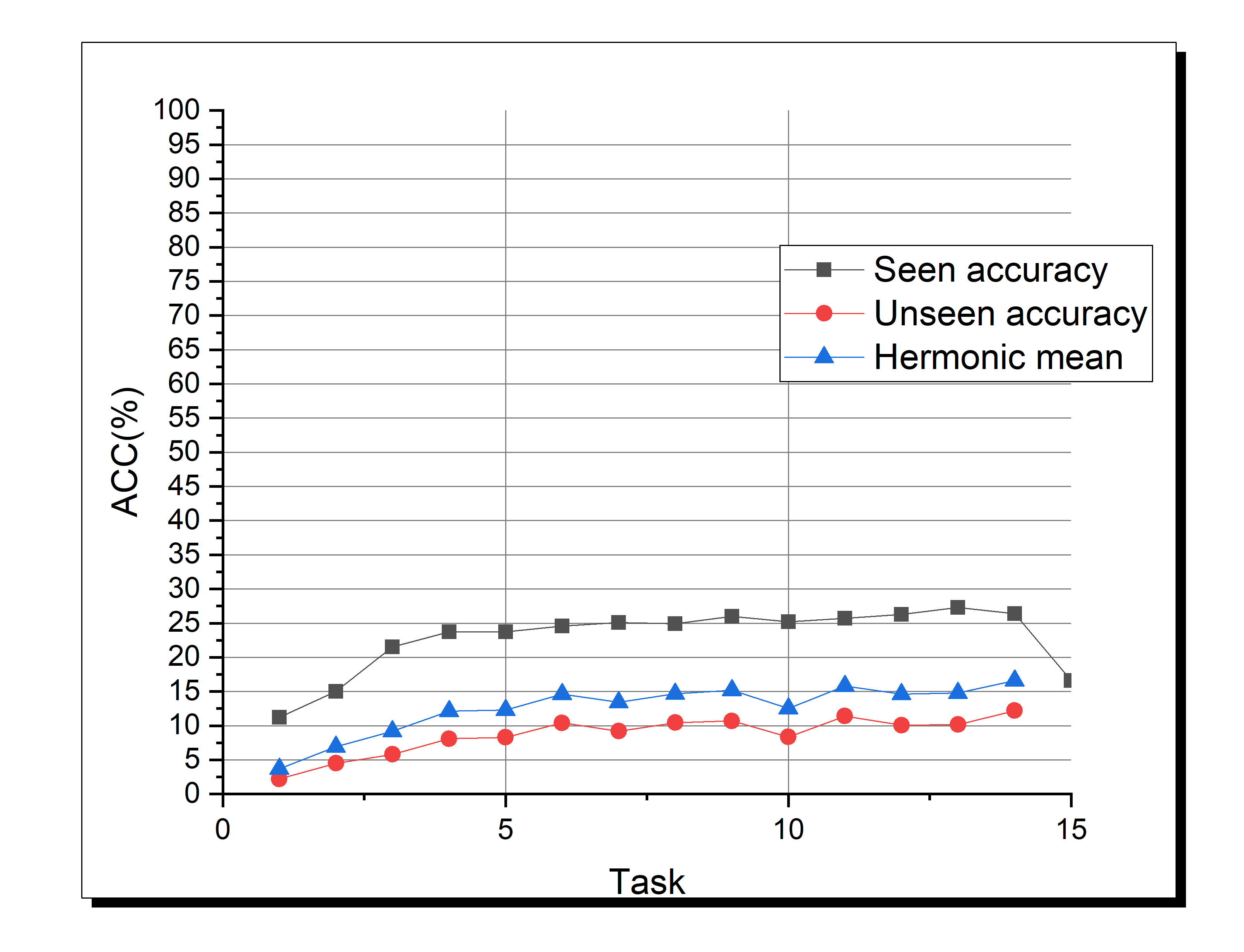}
	\caption{The SUN dataset results when we do not use the additional losses($L_e$ and $L_y$) during training the model.}
	\label{fig:f4}
\end{figure}

\begin{table}[hbt!]
	\centering
	\begin{tabular}{ |p{3cm}|p{1.2cm}|p{1.2cm}|p{1.2cm}|}
		\hline
		\multicolumn{4}{|c|}{SUN} \\
		\hline
		Methods & mSA & mUA & mH\\
		\hline
		AGEM+CZSL\cite{a10}   & -    &-&   10.50\\
		Seq-CVAE\cite{a7}&  16.88 & 11.40 & 13.38\\
		Seq-CADA\cite{a7} & 25.94 & 16.22 & 20.10\\
		\hline
		CZSL-CV+mof\cite{a7}& 23.21 & 13.20 & 16.71\\
		CZSL-CV+rb\cite{a7}&22.59 & 13.74 & 16.94\\
		CZSL-CV+res\cite{a7}&  23.99 & 14.10 & 17.63\\
		A-CZSL\cite{a9}$^{o}$& 17.2 & 6.31 & 9.68\\
		GRCZSL\cite{a8} & 17.74 & 11.50 & 13.73\\
		\hline
		\multicolumn{4}{|c|}{ours}\\
		\hline
		DVGR-CZSL$^{*}$ & 22.36&10.67 &14.54\\
		DVGR-CZSL& 22.9& 8.72 & 12.7\\
		\hline
		
	\end{tabular}
	\caption{Results for the SUN dataset, where mSA: Mean Seen Accuracy, mUA: Mean Unseen Accuracy, mH: Harmonic Mean Accuracy. ($^{o}$) denotes results are produced using original provided code in\cite{a9}. ($^{*}$) denotes $L_e$ and $L_y$ are used during training the model. The best results are presented over three runs.}
	\label{table:t2}
\end{table}
We divide the SUN dataset into 15 tasks, where each consists of around 47 classes. We compare our results with several methods in Table \ref{table:t2}. Generalized CZSL\cite{a7} is a memory-based approach, and it achieved mH = 17.3 though it has reported mH = 20.1 for Seq-CADA that has not been trained continually. GRCZSL\cite{a8} has used generative replay to alleviate catastrophic forgetting and achieved mH = 13.73; in contrast, our model achieves mH = 14.54 when each task gets trained for 101 epochs and the classifier for 25 epochs. The architecture consists of 13.9M parameters, and it takes around 3.5 hours to learn the SUN dataset when 50 synthetic samples per class are taken as the generative replay. The performance of our model per task is given in figures \ref{fig:f3} and \ref{fig:f4}.
\subsection{Performance on CUB Dataset}
\begin{figure}
	\includegraphics[width = 8cm, height = 5cm]{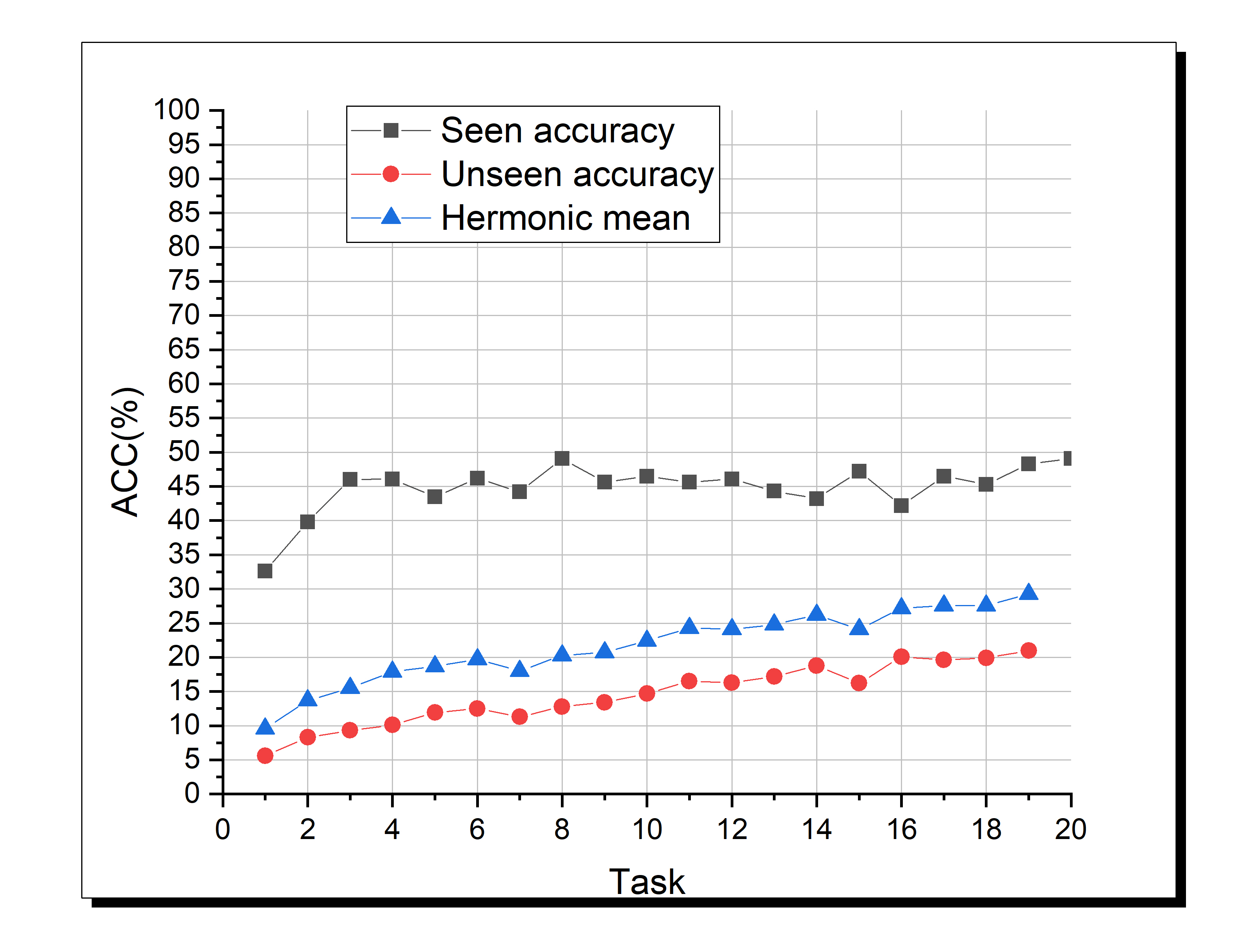}
	\caption{The CUB dataset results when we use the additional losses($L_e$ and $L_y$) during training the model.}
	\label{fig:f5}
	%    \vspace{<1>}
	\includegraphics[width = 8cm, height = 5cm]{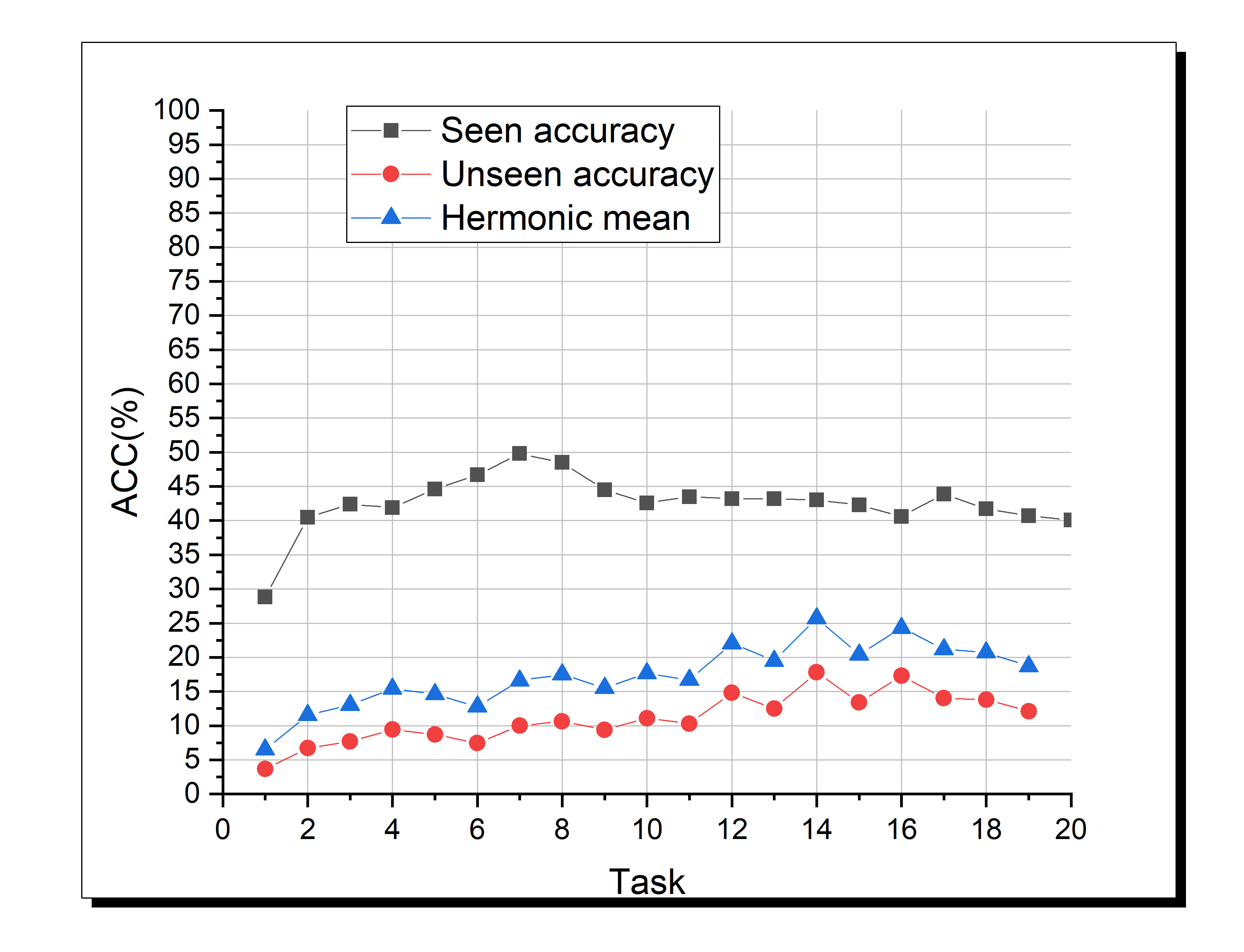}
	\caption{The CUB dataset results when we do not use the additional losses($L_e$ and $L_y$) during training the model.}
	\label{fig:f6}
\end{figure}

\begin{table}[hbt!]
	\centering
	\begin{tabular}{ |p{3cm}|p{1.2cm}|p{1.2cm}|p{1.2cm}|}
		\hline
		\multicolumn{4}{|c|}{CUB} \\
		\hline
		Methods & mSA & mUA & mH\\
		\hline
		AGEM+CZSL\cite{a10}   & -    &-&   13.20\\
		Seq-CVAE\cite{a7}&  24.66 & 8.57 & 12.18\\
		Seq-CADA\cite{a7} & 40.82 & 14.37 & 21.14\\
		\hline
		CZSL-CV+mof\cite{a7}& 43.73 & 10.26 & 16.34\\
		CZSL-CV+rb\cite{a7}&42.97 & 13.07 & 19.53\\
		CZSL-CV+res\cite{a7}&  44.89 & 13.45 & 20.15\\
		A-CZSL\cite{a9}& 34.25 & 12.42 & 17.41\\
		GRCZSL\cite{a8} & 41.91 & 14.12 & 20.48\\
		\hline
		\multicolumn{4}{|c|}{ours}\\
		\hline
		DVGR-CZSL$^{*}$ & 44.87&14.55 &21.66\\
		DVGR-CZSL & 42.84& 11.1 & 17.4\\
		\hline
	\end{tabular}
	\caption{Results for the CUB dataset, where mSA: Mean Seen Accuracy, mUA: Mean Unseen Accuracy, mH: Harmonic Mean Accuracy. ($^{*}$) denotes $L_e$ and $L_y$ are used during training the model. The best results are presented over three runs.}
	\label{table:t3}
\end{table}
We divide the CUB dataset into 20 tasks, where each consists of ten classes. We compare our results with several methods in Table \ref{table:t3}. Generalized CZSL\cite{a7} used memory to store real data as the replay, and it achieved mH = 20.15. GRCZSL\cite{a8} has used generative replay to alleviate catastrophic forgetting and achieved mH = 20.48, and A-CZSL\cite{a9} is an adversarial model that consists of two different modules and achieved mH = 17.41; in contrast, our model outperforms all existing models and achieves mH = 21.66 when each task gets trained for 101 epochs and the classifier for 25 epochs. The architecture consists of 171.8M parameters, and it takes around 1.6 hours to learn the CUB dataset when 50 synthetic samples per class are taken as the generative replay. The performance of our model per task is given in figures \ref{fig:f5} and \ref{fig:f6}.
\begin{table*}
	\centering
	\begin{tabular}{ |p{1.1cm}|p{2.1cm}|p{1.3cm}|p{1.2cm}|p{1.3cm}|p{1.2cm}|p{1.2cm}|p{1.2cm}|p{1.4cm}|p{1.2cm}|}
		\hline
		Dataset & AGEM+CZSL&Seq-CVAE&Seq-CADA&CZSL-CV+mof&CZSL-CV+rb&CZSL-CV+res&A-CZSL&GRCZSL&DVGR-CZSL\\
		\hline
		SUN & 10.5& 13.38&20.10&16.71&16.94&17.63&9.68&13.73&14.54\\
		\hline
		CUB &13.20&12.18&21.14&16.34&19.53&20.15&17.41&20.48&21.66\\
		\hline
		aPY &16.30&18.33&16.42&18.27&18.6&23.9&23.05&20.12&31.7\\
		\hline
		AWA1&25.73&27.14&27.59&30.46&33.64&35.51&37.19&33.56&38\\
		\hline
		AWA2&22.24&28.67&30.38&36.6&37.32&38.34&37.19&35.82&40.6\\
		\hline
	\end{tabular}
	\caption{compares of Harmonic mean accuracy of different methods for SUN, CUB, aPY, AWA1, and AWA2 datasets.}
	\label{table:t7}
\end{table*}
\subsection{Performance on aPY Dataset}
\begin{figure}
	\includegraphics[width = 8cm, height = 5cm]{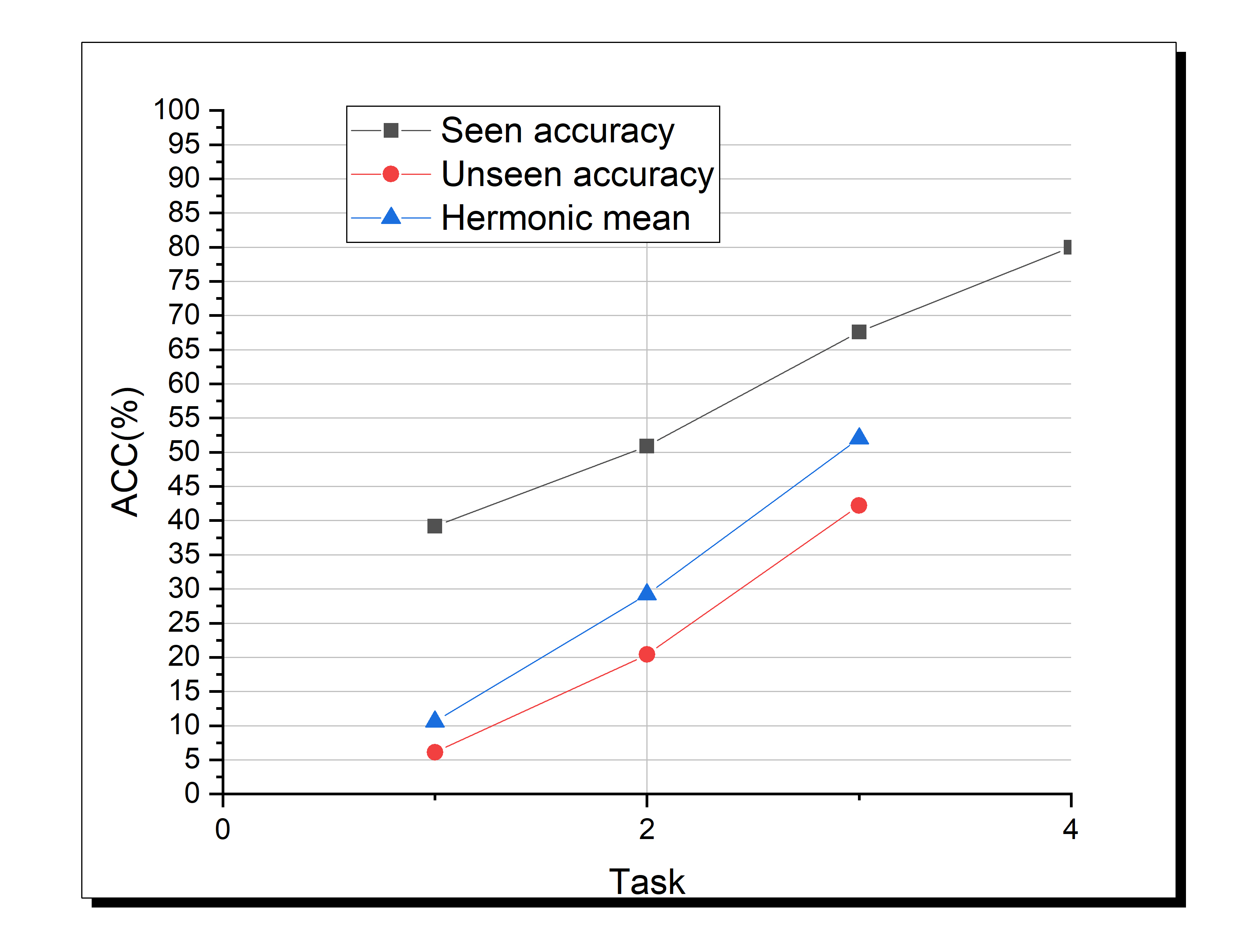}
	\caption {The aPY dataset results when we use the additional losses($L_e$ and $L_y$) during training the model.}
	\label{fig:f7}
	%    \vspace{<1>}
	%\includegraphics[width = 8cm, height = 5cm]{aPY_s1_wl.jpg}
	%\caption{The aPY dataset results when we do not use the additional losses($L_e$ and $L_y$) during training the model.}
	%\label{fig:f8}
\end{figure}

\begin{table}[hbt!]
	\centering
	\begin{tabular}{ |p{3cm}|p{1.2cm}|p{1.2cm}|p{1.2cm}|}
		\hline
		\multicolumn{4}{|c|}{aPY} \\
		\hline
		Methods & mSA & mUA & mH\\
		\hline
		AGEM+CZSL\cite{a10}$^{**}$   & -    &-&   16.30\\
		Seq-CVAE\cite{a7}&  51.57 & 11.8 & 18.33\\
		Seq-CADA\cite{a7} & 45.25 & 10.59 & 16.42\\
		\hline
		CZSL-CV+mof\cite{a7}& 64.91 & 10.79 & 18.27\\
		CZSL-CV+rb\cite{a7}&64.45 & 11.00 & 18.60\\
		CZSL-CV+res\cite{a7}&  64.88 & 15.24 & 23.90\\
		A-CZSL\cite{a9}& 58.14 & 15.91 & 23.05\\
		GRCZSL\cite{a8} & 66.47 & 12.06 & 20.12\\
		\hline
		\multicolumn{4}{|c|}{ours}\\
		\hline
		DVGR-CZSL$^{*}$ & 59.4&22.9 &30.55\\
		DVGR-CZSL & 62.5 & 22.7 & 31.7\\
		\hline
	\end{tabular}
	\caption{Results for the aPY dataset, where mSA: Mean Seen Accuracy, mUA: Mean Unseen Accuracy, mH: Harmonic Mean Accuracy. ($^{*}$) denotes $L_e$ and $L_y$ are used during training the model. ($^{**}$) denotes the result has been produced using the original provided code in\cite{a10}. The best results are presented over three runs.}
	\label{table:t4}
\end{table}
We divide the aPY dataset into four tasks, where each consists of eight classes. We compare our results with several models in Table \ref{table:t4}. Generalized CZSL\cite{a7} used memory to store real data as the replay to alleviate catastrophic forgetting, and it achieved mH = 25.84. GRCZSL\cite{a8} has used generative replay and achieved mH = 20.12, and A-CZSL\cite{a9} is an adversarial model that achieved mH = 23.05; in contrast, our model outperforms all existing models and achieves mH = 31.7 when each task gets trained for 101 epochs and the classifier for 25 epochs. The architecture consists of 3.1M parameters, and it takes around 220 seconds to learn the aPY dataset when 50 synthetic samples per class are taken as the generative replay. The performance of our model per task is given in figures \ref{fig:f7}.. 
\subsection{Performance on AWA1 Dataset}
\begin{figure}
	\centering
	\includegraphics[width = 8cm, height = 5cm]{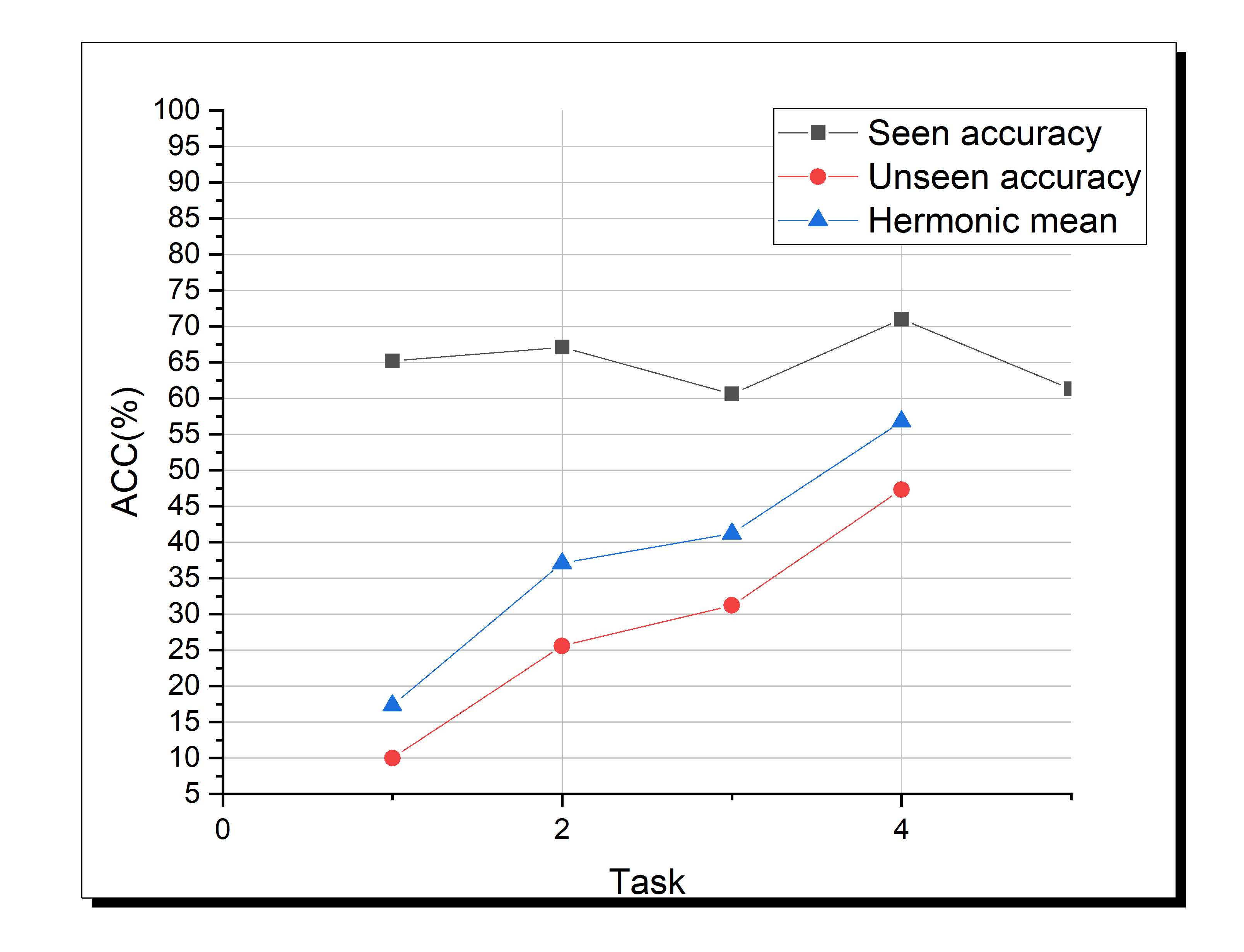}
	\caption{The AWA1 dataset results when we use the additional losses($L_e$ and $L_y$) during training the model.}
	\label{fig:f9}
	%    \vspace{<1>}
	%\includegraphics[width = 8cm, height = 5cm]{AWA2_s1_wl.jpg}
	%\caption{The AWA1 dataset results when we do not use the additional losses($L_e$ and $L_y$) during training the model.}
	%\label{fig:f10}
\end{figure}

\begin{table}[hbt!]
	\centering
	\begin{tabular}{ |p{3cm}|p{1.2cm}|p{1.2cm}|p{1.2cm}|}
		\hline
		\multicolumn{4}{|c|}{AWA1}\\
		\hline
		Methods & mSA & mUA & mH\\
		\hline
		AGEM+CZSL\cite{a10}$^{**}$   & -    &-&   25.73\\
		Seq-CVAE\cite{a7}&  59.27 & 18.24 & 27.14\\
		Seq-CADA\cite{a7} & 51.57 & 18.02 & 27.59\\
		\hline
		CZSL-CV+mof\cite{a7}& 76.77 & 19.26 & 30.46\\
		CZSL-CV+rb\cite{a7}&77.85 & 21.90 & 33.64\\
		CZSL-CV+res\cite{a7}&  78.56 & 23.65 & 35.51\\
		A-CZSL\cite{a9}& 70.16 & 25.93 & 37.19\\
		GRCZSL\cite{a8} & 78.66 & 21.86 & 33.56\\
		\hline
		\multicolumn{4}{|c|}{ours}\\
		\hline
		DVGR-CZSL$^{*}$ & 65.1&28.5 &38\\
		DVGR-CZSL & 68.4 & 26.9 & 37.2\\
		\hline
	\end{tabular}
	\caption{Results for the AWA1 dataset, where mSA: Mean Seen Accuracy, mUA: Mean Unseen Accuracy, mH: Harmonic Mean Accuracy. ($^{*}$) denotes $L_e$ and $L_y$ are used during training the model. ($^{**}$) denotes the result has been produced using the original provided code in\cite{a10}. The best results are presented over three runs.}
	\label{table:t5}
\end{table}
We divide the AWA1 dataset into five tasks, where each consists of ten classes. We compare our results with several methods in Table \ref{table:t5}. Generalized CZSL\cite{a7} used memory replay to alleviate catastrophic forgetting, and it achieved mH = 35.51. GRCZSL\cite{a8} has used generative replay and achieved mH = 33.56, and A-CZSL\cite{a9} is an adversarial model that achieved mH = 35.75; in contrast, our model outperforms all existing models and achieves mH = 38 when each task gets trained for 101 epochs and the classifier for 25 epochs. The architecture consists of 3.9M parameters, and it takes around 450 seconds to learn the AWA1 dataset when 50 synthetic samples per class are taken as the generative replay. The performance of our model per task is given in figures \ref{fig:f9}.
\subsection{Performance on AWA2 Dataset}
\begin{figure}
	\centering
	\includegraphics[width = 8cm, height = 5cm]{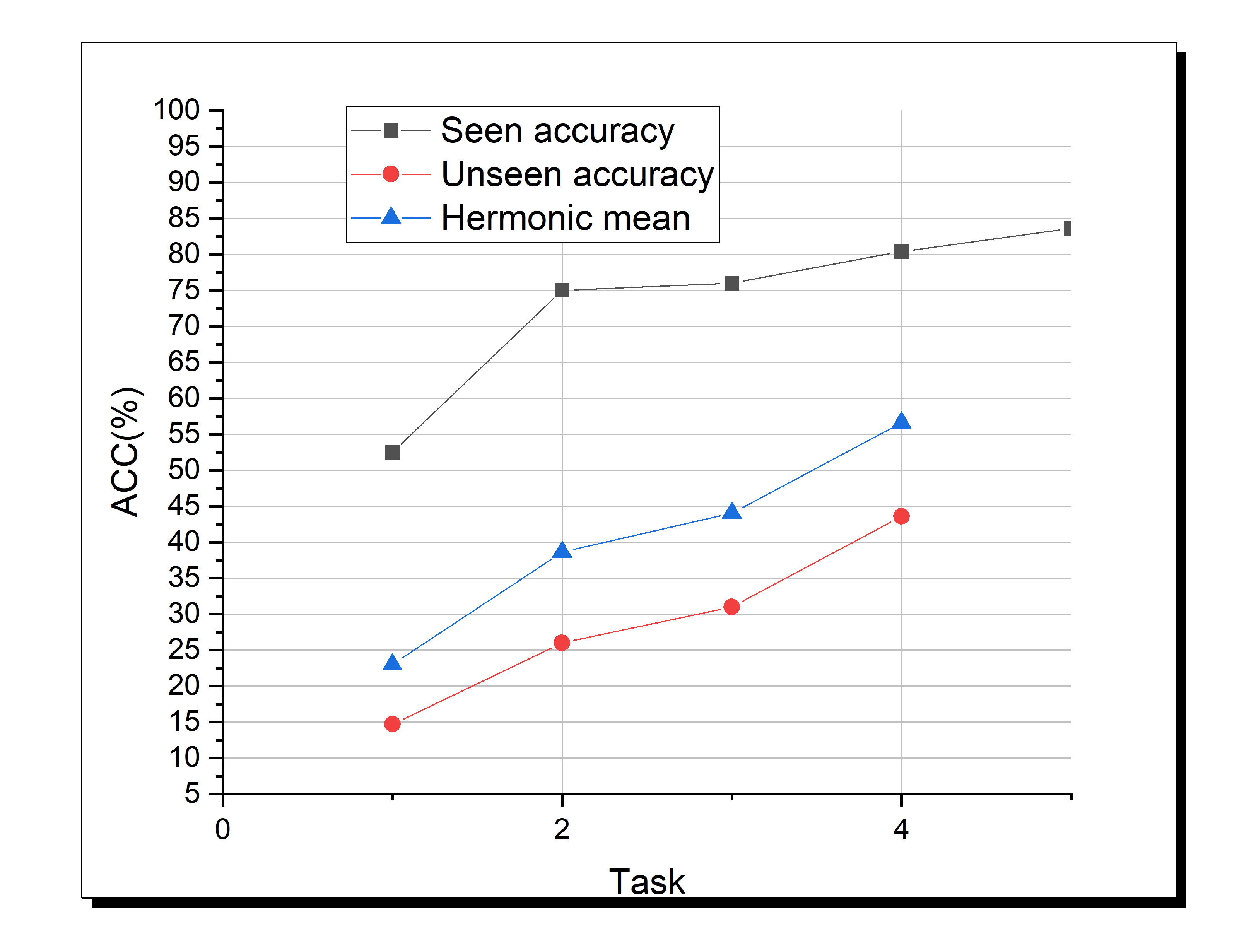}
	\caption{The AWA2 dataset results when we use the additional losses($L_e$ and $L_y$) during training the model.}
	\label{fig:f11}
	%    \vspace{<1>}
	%\includegraphics[width = 8cm, height = 5cm]{AWA2_s1_wl.jpg}
	%\caption{The AWA2 dataset results when we do not use the additional losses($L_e$ and $L_y$) during training the model.}
	%\label{fig:f12}
\end{figure}

\begin{table}
	\centering
	\begin{tabular}{ |p{3cm}|p{1.2cm}|p{1.2cm}|p{1.2cm}|}
		\hline
		\multicolumn{4}{|c|}{AWA2} \\
		\hline
		Methods & mSA & mUA & mH\\
		\hline
		AGEM+CZSL\cite{a10}$^{**}$   & -    &-&   27.24\\
		Seq-CVAE\cite{a7}&  61.42 & 19.34 & 28.67\\
		Seq-CADA\cite{a7} & 52.30 & 20.30 & 30.38\\
		\hline
		CZSL-CV+mof\cite{a7}& 79.11 & 24.41 & 36.6\\
		CZSL-CV+rb\cite{a7}&80.92 & 24.82 & 37.32\\
		CZSL-CV+res\cite{a7}&  80.97 & 25.75 & 38.34\\
		A-CZSL\cite{a9}& 70.16 & 25.93 & 37.19\\
		GRCZSL\cite{a8} & 81.01 & 23.59 & 35.82\\
		\hline
		\multicolumn{4}{|c|}{ours}\\
		\hline
		DVGR-CZSL$^{*}$ & 73.5&28.8 &40.6\\
		DVGR-CZSL & 73.4 & 27.05 & 38.3\\
		\hline
	\end{tabular}
	\caption{Results for the AWA2 dataset, where mSA: Mean Seen Accuracy, mUA: Mean Unseen Accuracy, mH: Harmonic Mean Accuracy. ($^{*}$) denotes $L_e$ and $L_y$ are used during training the model. ($^{**}$) denotes the result has been produced using the original provided code in\cite{a10}. The best results are presented over three runs.}
	\label{table:t6}
\end{table}
We divide the AWA2 dataset into five tasks, where each consists of ten classes. We compare our results with several methods in Table \ref{table:t6}. Generalized CZSL\cite{a7} used memory replay to alleviate catastrophic forgetting, and it achieved mH = 38.34. GRCZSL\cite{a8} has used generative replay and achieved mH = 35.82, and A-CZSL\cite{a9} is an adversarial model that achieved mH = 37.19; in contrast, our model outperforms all existing models and achieves mH = 40.6 when each task gets trained for 101 epochs and the classifier for 25 epochs. The architecture consists of 3.9M parameters, and it takes around 800 seconds to learn the AWA2 dataset when 50 synthetic samples per class are taken as a generative replay. The performance of our model per task is given in figures \ref{fig:f11}.

\subsection{Few Things We Tried}
\begin{figure}[hbt!]
	\centering
	\includegraphics[width = 8cm, height = 5cm]{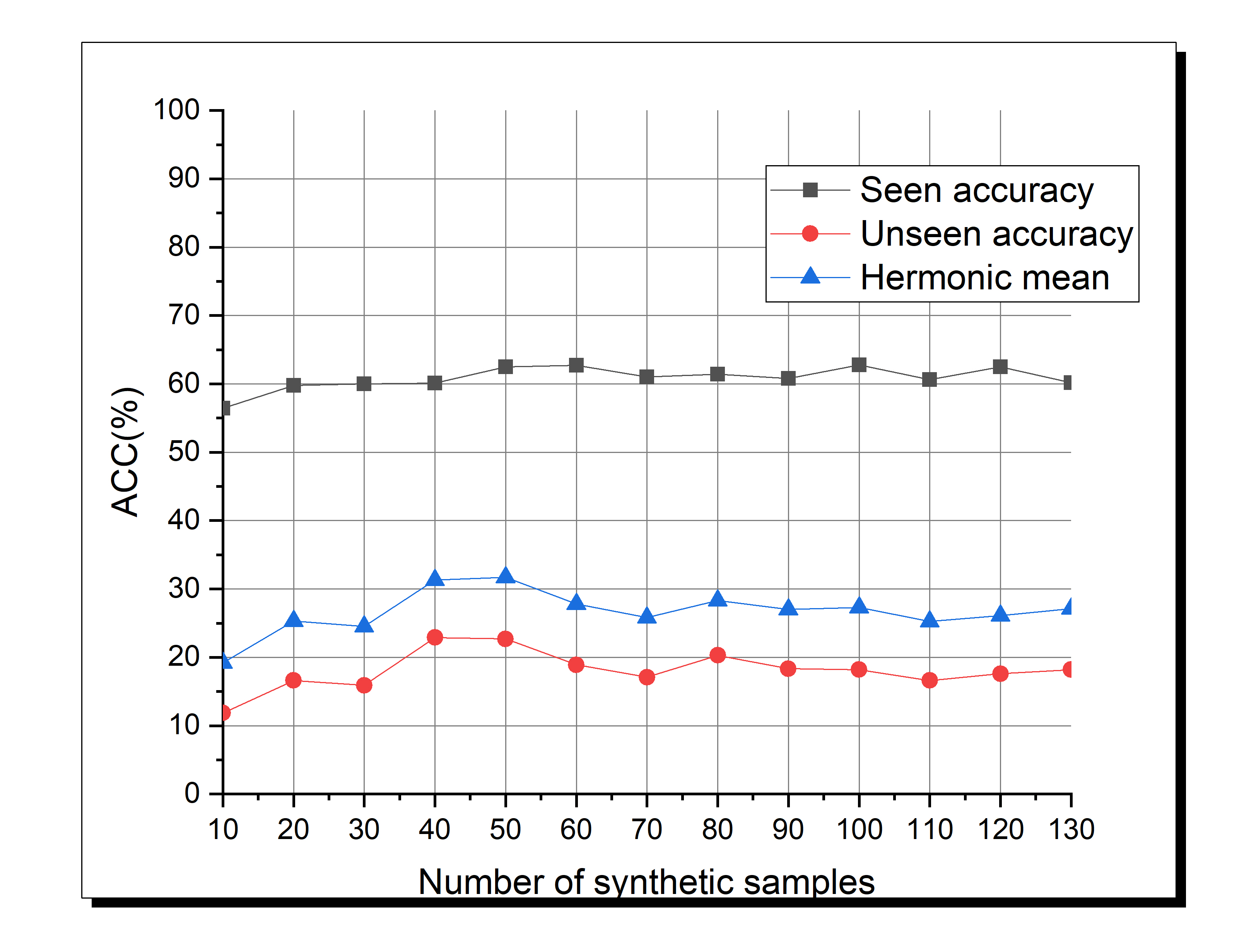}
	\caption{The aPY dataset results using a different number of synthetic samples per class as the generative replay. Here, we do not use the additional losses($L_e$ and $L_y$) during training the model.}
	\label{figure:f13}
\end{figure}

\begin{itemize}
	\item First, we added skip connections, sent the information from encoders to decoders, but got drastic deterioration in performance.
	\item Second, if we do not use rectified linear units(ReLU) function in the classifier, we would get around 5-10\% decay in harmonic mean accuracy depending on the dataset. 
	\item Third, the number of synthetic samples per class as the generative replay is optimum between 40-50. Figure \ref{figure:f13} shows that harmonic mean accuracy is maximum at that range for the aPY dataset.
\end{itemize}
\section{Conclusion}
In this work, we proposed a novel algorithm for continual zero-shot learning that takes structure-based and generative replay-based approaches together into account to fight against catastrophic forgetting. The model's performance is evaluated using five ZSL benchmark datasets, out of which we achieve state-of-the-art results for four datasets. We can apply this approach for object detection in future work. What should be the optimum latent dimension, hidden-layer size?
\section{Acknowledgement}
    We thank \href{https://ece.iisc.ac.in/~vinod/}{Prof. Vinod Sharma} and \href{https://ece.iisc.ac.in/~parimal/}{Prof. Parimal Parag} from the Department of ECE, Indian Institute of Science(IISc), Bangalore, for the course Foundation of Machine Learning that helped me a lot to understand basics of Machine Learning. I started my journey in Machine Learning since that course. 
\clearpage
\newpage
{\small
\bibliographystyle{plain}
\bibliography{AA}
}

\end{document}